# Un système multi-agents d'aide à la décision clinique fondé sur des ontologies

SHEN Ying [1], JACQUET-ANDRIEU Armelle [2], COLLOC Joël [3]

Université Paris Ouest / CNRS, UMR 7114, MoDyCo[1,2]
200 avenue de la République, 92000 NANTERRE [1,2]
France[1,2]
ying.shen@u-paris10.fr[1]
armelle.jacquet@u-paris10.fr[2]

Université du Havre CIRTAI IDEES UMR CNRS 6266[3]
25, rue Philippe Lebon 76086 Le HAVRE Cedex[3]
France[3]
joel.colloc@univ-lehavre.fr[3]

**RESUME**

Les systèmes d'aide à la décision cliniques combinent des connaissances et des données provenant de sources diverses, représentées à l'aide de modèles quantitatifs fondés sur des méthodes stochastiques, ou qualitatifs reposant plutôt sur les heuristiques d'experts et le raisonnement déductif. Parallèlement, le raisonnement à partir de cas (RàPC) mémorise et restitue l'expérience de résolution de problèmes similaires. La coopération de bases de connaissances cliniques hétérogènes (objets connaissances, distances sémantiques, fonctions d'évaluation, règles logiques, bases de données...) repose sur les ontologies médicales. Un système multi-agents d'aide à la décision (SMAAD) permet l'intégration et la coopération des agents spécialisés dans différents domaines de connaissances (sémiologie, pharmacologie, cas cliniques, etc.). Chaque agent spécialisé exploite une base de connaissances définissant les conduites à tenir conformes à l'état de l'art associé à une base ontologique qui exprime les relations sémantiques entre les termes du domaine considéré. Notre approche repose sur la spécialisation d'agents adaptés aux modèles de connaissances utilisés lors des étapes de la démarche clinique et les ontologies. Cette approche modulaire est adaptée à la réalisation de SMAAD dans de nombreux domaines.

## 1. Introduction

Un système multi-agent d'aide à la décision (SMAAD) utilise des bases de données, des bases connaissances, des ontologies, des modes de raisonnements variés qu'il combine selon la démarche clinique. On distingue les modèles quantitatifs, s'appuyant sur des méthodes stochastiques, utilisés dans les études épidémiologiques, et les modèles qualitatifs qui reposent sur les connaissances heuristiques d'experts et le raisonnement déductif. Nous utilisons un système multi-agent supervisé, afin d'orchestrer les différentes étapes de la démarche clinique et aboutir à une solution correcte et efficace en un temps limité. Toutefois, les solutions émergent de la coopération d'agents spécialisés dans les étapes cliniques (diagnostic, pronostic, thérapeutique, suivi thérapeutique) et des bases de connaissances élaborées avec des modèles différents. Par exemple, le raisonnement à partir de cas (RàPC) est un mode de résolution analogique des problèmes qui compare de nouveaux cas à partir de cas précédents indexés [1] Ainsi, un système d'aide au traitement du diabète combine (RàPC), des règles de production et des statistiques.

Notre approche s'appuie sur un type d'agent clinique général (figure 2). Il constitue la racine d'une hiérarchie de types d'agents cliniques spécialisés d'abord selon les modèles, puis selon les domaines de connaissances (figure 4). Cet article montre l'apport des ontologies à la coopération de bases de connaissances dans un SMAAD. L'approche proposée est comparée à d'autres. Un exemple est présenté en annexe.

## 2. Modèle de la démarche clinique

Le système multi-agent (SMA) que nous proposons s'appuie sur nos précédents travaux [6,19] et ceux d'autres auteurs [14,18]. Dans les SMA non-supervisées, la communication entre agents a lieu à l'aide de langages comme ACL ou KQML [9].

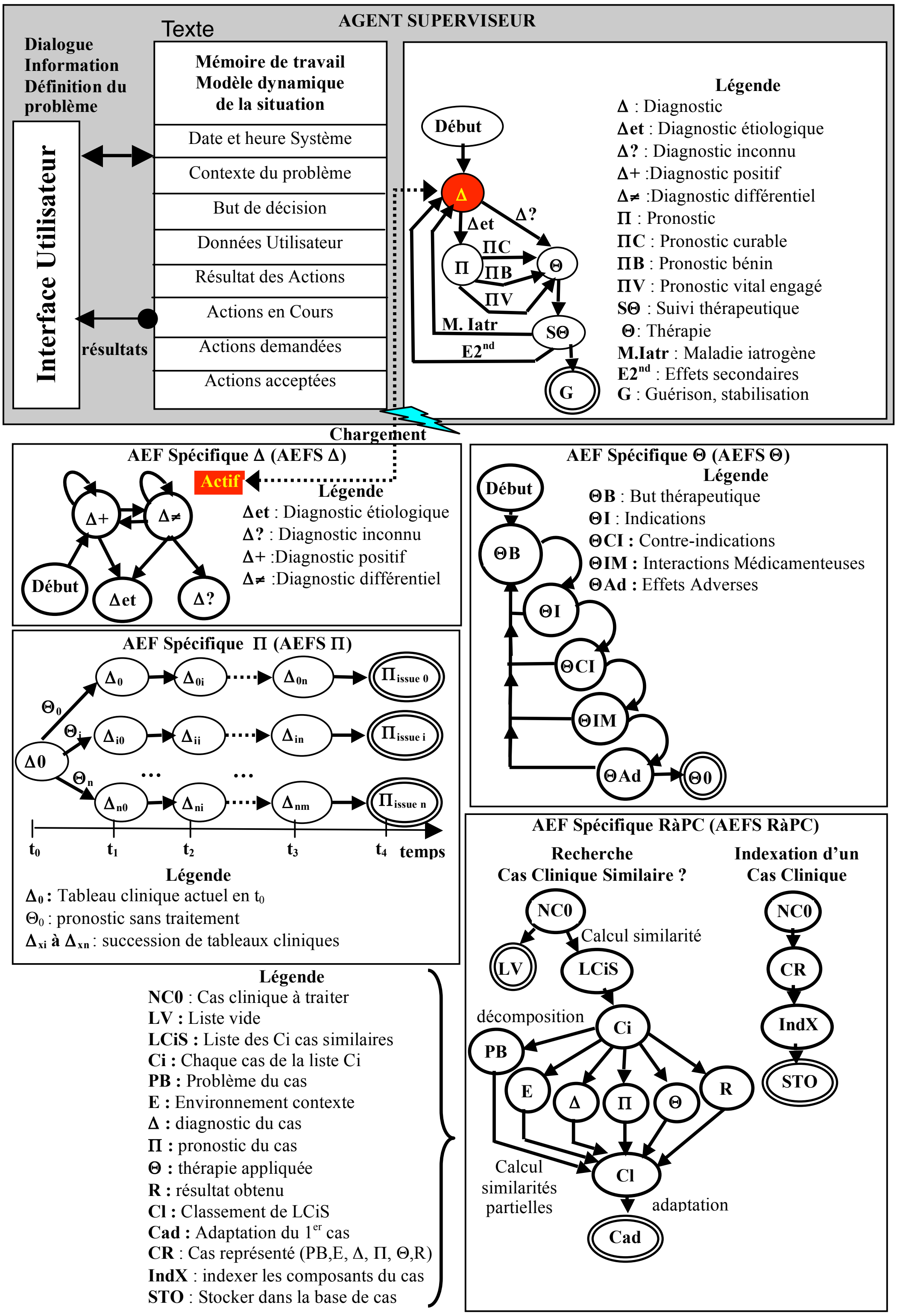

**Figure 1. Agent superviseur, AEF clinique général et AEF spécifiques Δ, Π, Θ et RàPC**

Dans un SMA supervisé, l'agent superviseur gère la communauté des agents (figure 1).

La cohérence des transactions est assurée par des automates d'états finis **(AEF)** chargés de planifier et d'organiser les tâches décisionnelles. Le superviseur utilise un Automate d'Etat Fini Clinique Général (**AEFCG**) définissant l'ensemble des étapes nécessaires à la démarche de décision clinique qui charge et déclenche les automates d'état finis contrôlant des tâches décisionnelles spécifiques AEFS Δ,Π,Θ,RàPC (figure 1). L'AEFS RàPC offre un mode particulier de consultation et de restitution des cas cliniques traités. Il constitue la mémoire et l'expérience du SMAAD. Ces AEFS sont chargés à partir d'une base (figure 1 et 4). La mémoire de travail décrit la situation courante et recueille les résultats intermédiaires. L'AEFCG ordonnance l'ensemble du processus de décision. Chaque AEFS gère le déroulement d'une étape clinique spécifique diagnostic Δ, pronostic Π, thérapie Θ, suivi thérapeutique SΘ (non représenté par concision). Ces agents encapsulent des modèles de connaissances et de décision différents. Le superviseur assure le dialogue avec l'utilisateur. Il contrôle la prise en charge et l'exécution des tâches cliniques par les agents disponibles compétents. Un agent doit abandonner une tâche si le contexte ou l'objectif actuel fixé par le superviseur est modifié : une solution trouvée par un ou plusieurs autres agents a été validée par l'utilisateur, le délai imparti a expiré. Chaque agent utilise ses connaissances réflexives (figure 2) pour déterminer s'il peut participer à la tâche demandée par le superviseur. Son degré de spécialisation, la nature de la tâche, les connaissances disponibles dans la base, le mode de raisonnement interviennent dans ce choix. Si l'agent accepte, le superviseur l'inscrit dans les actions en cours. L'agent devient actif. Il consulte les données de l'utilisateur dans la mémoire du superviseur, il sollicite des informations via l'interface utilisateur. Les réponses sont classées et disponibles dans la mémoire de travail du superviseur pour tous les agents activés par la tâche courante. Lorsque l'objectif est atteint, le superviseur charge l'AEFS de l'étape suivante conformément aux transitions de l'AEFCG (figure 1).
Notre approche définit de manière simple, robuste et modulaire, d'abord, l'ensemble du protocole décisionnel clinique par un AEFCG puis chaque tâche spécifique par un AEFS Δ, Π, Θ, SΘ [19]. Le développement de nouveaux AEFCG et AEFS facilite l'adaptation du SMAAD à de nombreux domaines où un protocole clinique est pertinent. Nous avons montré dans de précédents travaux [6, 19] que l'émergence de décisions utiles à un SMAAD clinique doit être supervisé par la démarche clinique afin d'obtenir un résultat en temps utile. D'une part, le dialogue avec l'utilisateur doit avoir lieu *via* une interface (figure1) ; d'autre part, la gestion de la cohérence dans un SMA supervisé est plus facile. Les inconvénients proviennent de la surcharge de l'interface du superviseur et d'une relative diminution de l'autonomie des agents. La figure 2 décrit le Type d'Agent Clinique Général (**TACG**) dont les modules sont fonctionnellement liés les uns aux autres et assurent la communication, la négociation, la décision, l'évaluation du contexte (exprimé par le superviseur) et la gestion des tâches.

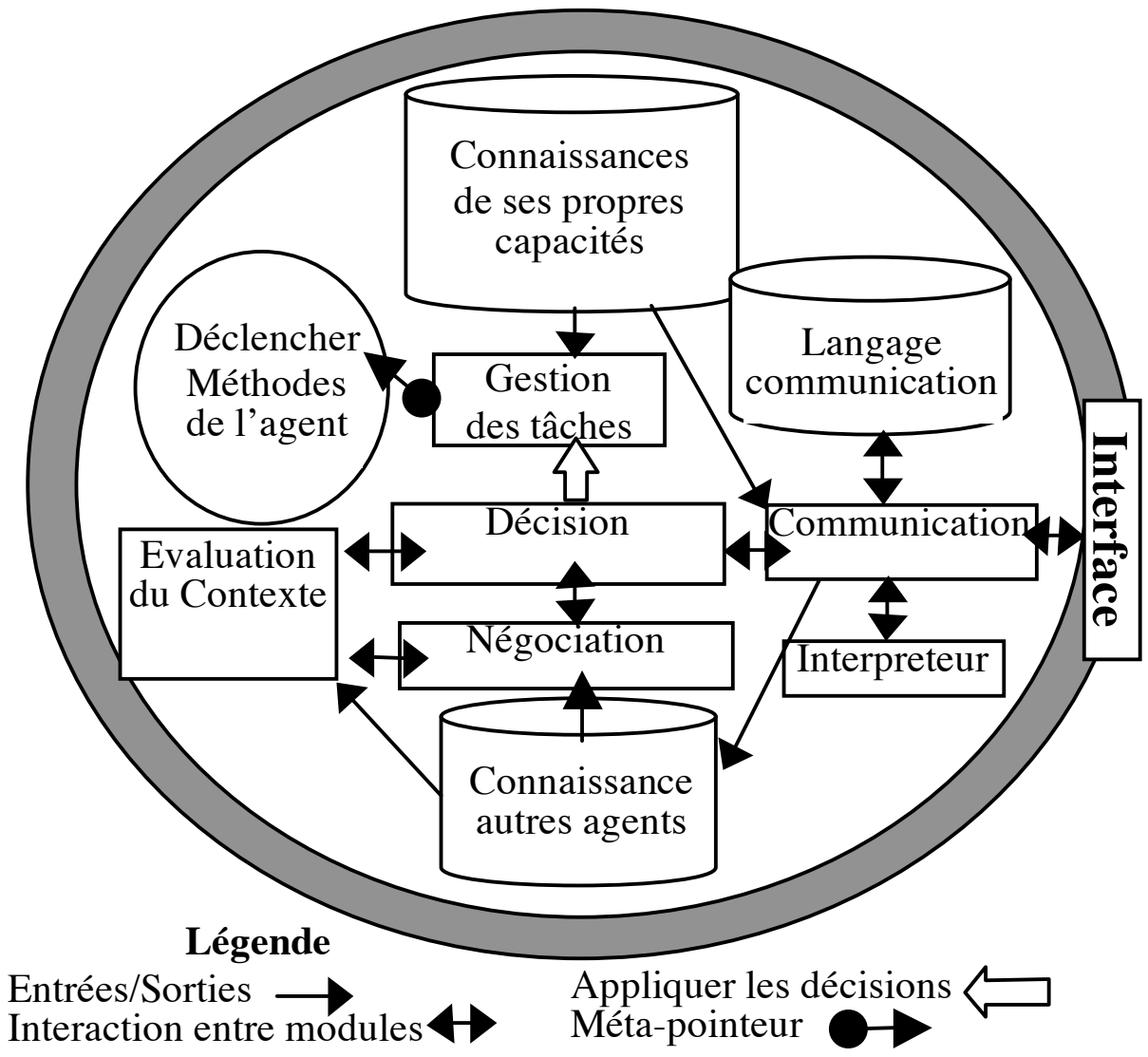

**Figure 2. Architecture du Type d'Agent Clinique Général (TACG)**

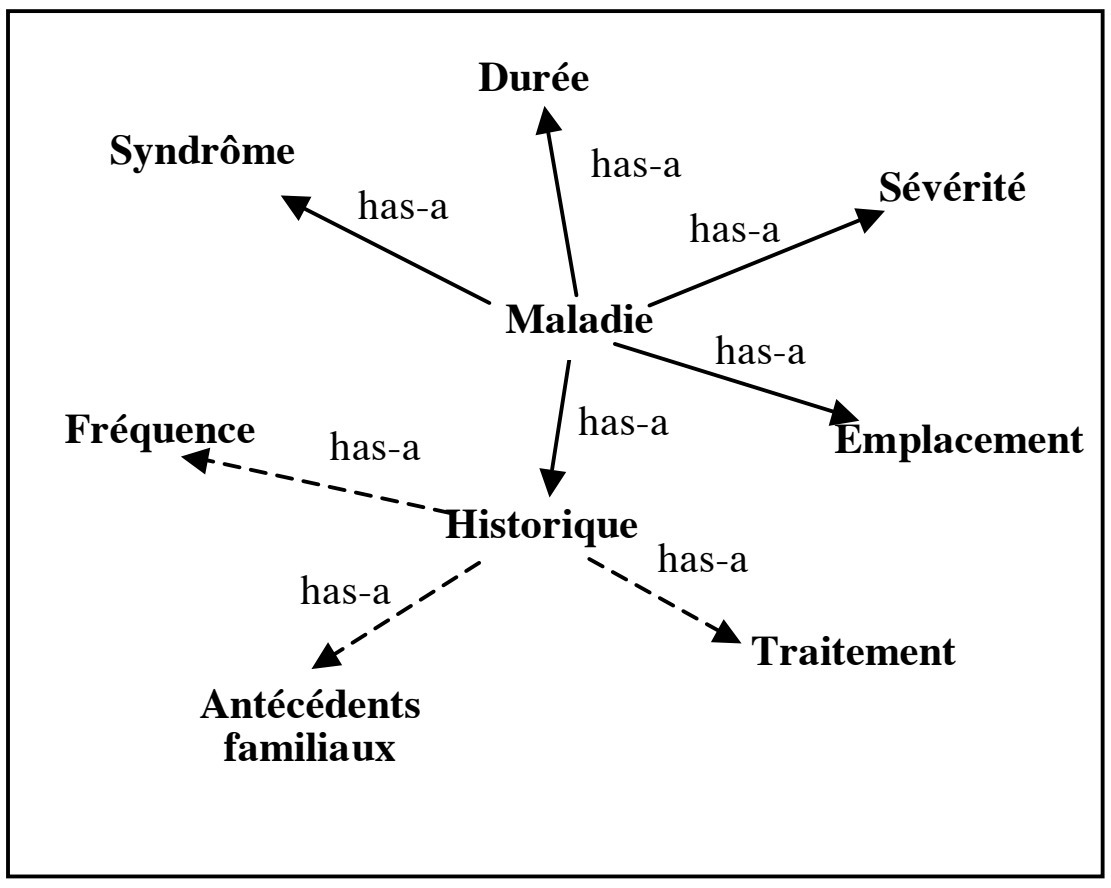

**Figure 3. Hiérarchisation de l'ontologie du diagnostic d'une maladie : la classe historique**

## 3. Apport des ontologies

Les modèles issus des travaux de linguistes comme Quillian, sur la sémantique, ont été repris et adaptés par les informaticiens. Les modèles d'ontologie, fondés sur des graphes, sont destinés à l'axiomatisation d'un domaine de connaissances (figure 3). Ils peuvent être implantés à l'aide de la logique formelle, de modèles à règles de production, de langages sémantiques comme XML et ses dérivés OWL, de langages fonctionnels comme Objective Caml.

### 3.1. Modèles d'ontologies

L'élaboration d'une ontologie se décompose en deux étapes : expliquer les termes utilisés pour décrire le déroulement clinique habituel, puis établir les liens d'interactions entre le Dictionnaire Ontologique et le Module de Communication. Le développement d'une ontologie comprend les étapes suivantes : définition, implémentation, intégration et documentation. La définition concerne l'identification des concepts élémentaires et leurs relations dans le domaine considéré, notamment, la spécialisation de classes en sous-classes [8]. L'ontologie diagnostic (figure 3) s'appuie sur des taxonomies médicales et les ontologies existantes : le réseau sémantique UMLS [2,3], le *Medical Subject Headings* (MeSH) [3] et le thésaurus NCI des maladies infectieuses. D'autres relations sémantiques lient les entités maladies aux hypothèses de diagnostic médical. Le réseau sémantique UMLS [2,3] inclut les relations suivantes indiquant les types de facteurs pris en compte durant le diagnostic : is_a (est-un), has_a (a une propriété), associated_with (est associé_à) physically_related_to (lié_physiquement_à), functionally_related_to (lié_fonctionnellement_à), temporally_related_to (lié_temporellement_à), conceptually_related_to (lié_conceptuellement_à). Chaque étiquette représente une classe ou un concept et chaque flèche une relation. Un élément inclus dans le diagnostic de la maladie correspond à des pistes de diagnostic applicables à un groupe de patients ayant une maladie similaire.

Le Dictionnaire Ontologique et le Module de Communication expriment les interactions entre la terminologie liée aux domaines et les bases de connaissances cliniques. L'interface exploite les liens sémantiques OWL pour exprimer les termes et concepts utilisés par les praticiens ou les infirmières dans le cadre de leur activité. Des relations d'extrapolations combinent le *Semantic Web Rule Language* (SWRL) et OWL, issu de XML, pour établir de nouvelles relations sans modifier celles existantes [8]. Le modèle clinique est structuré par une hiérarchie de concepts complexes représentés comme une combinaison de sous-concepts unis entre eux par des relations de spécialisation ou de composition à différents niveaux de granularité (figures 3.1 à 3.3).

### 3.2. Notre modèle ontologique

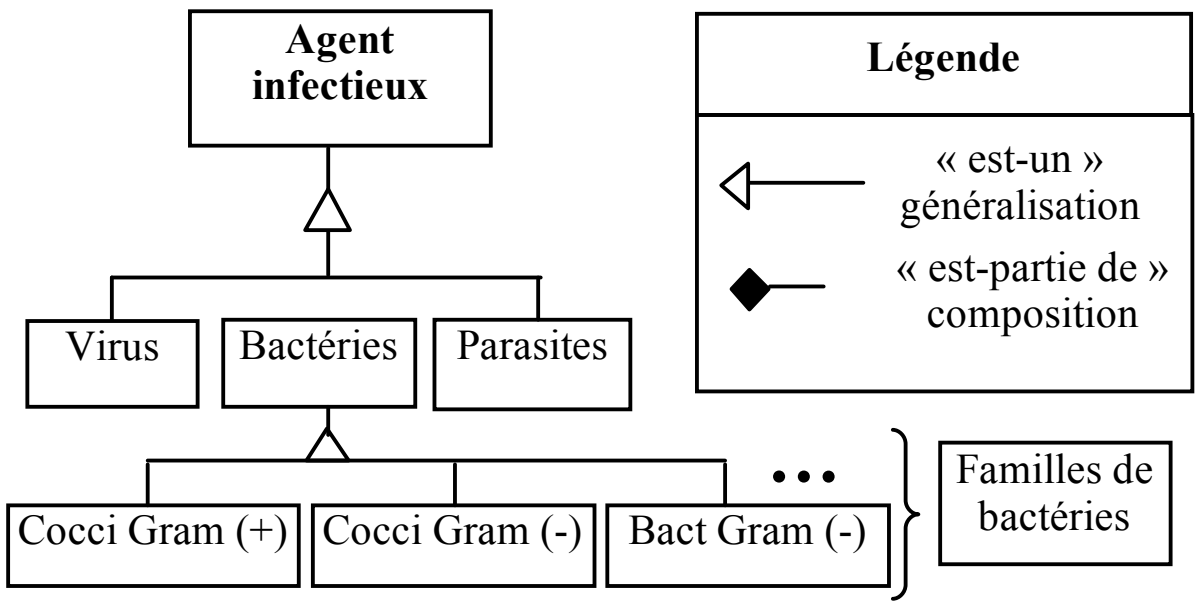

**Figure 3.1 Ontologie des agents infectieux**

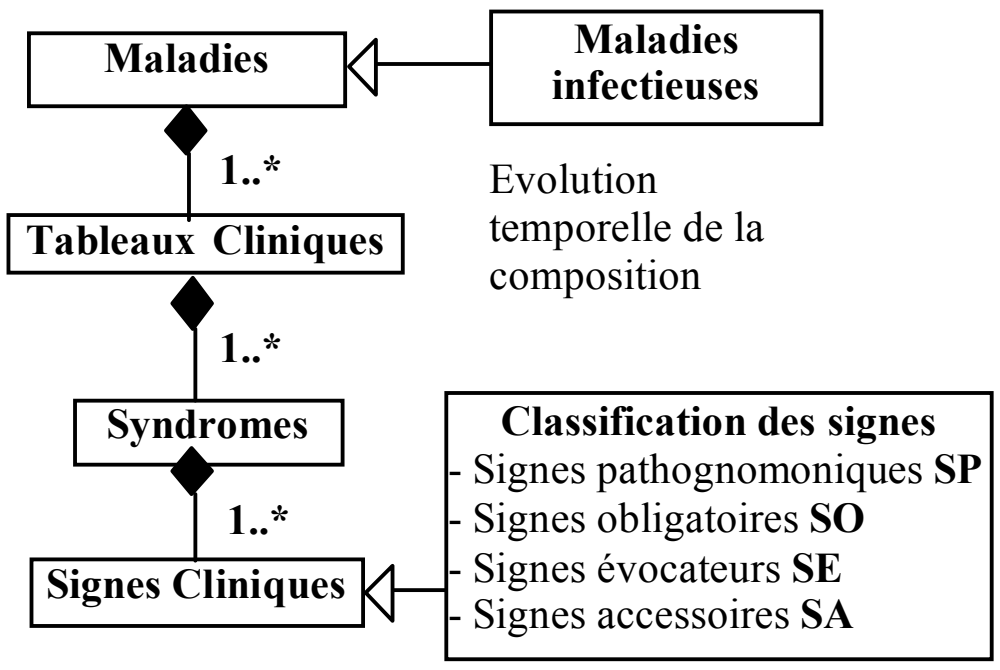

**Figure 3.2 Ontologies des pathologies**

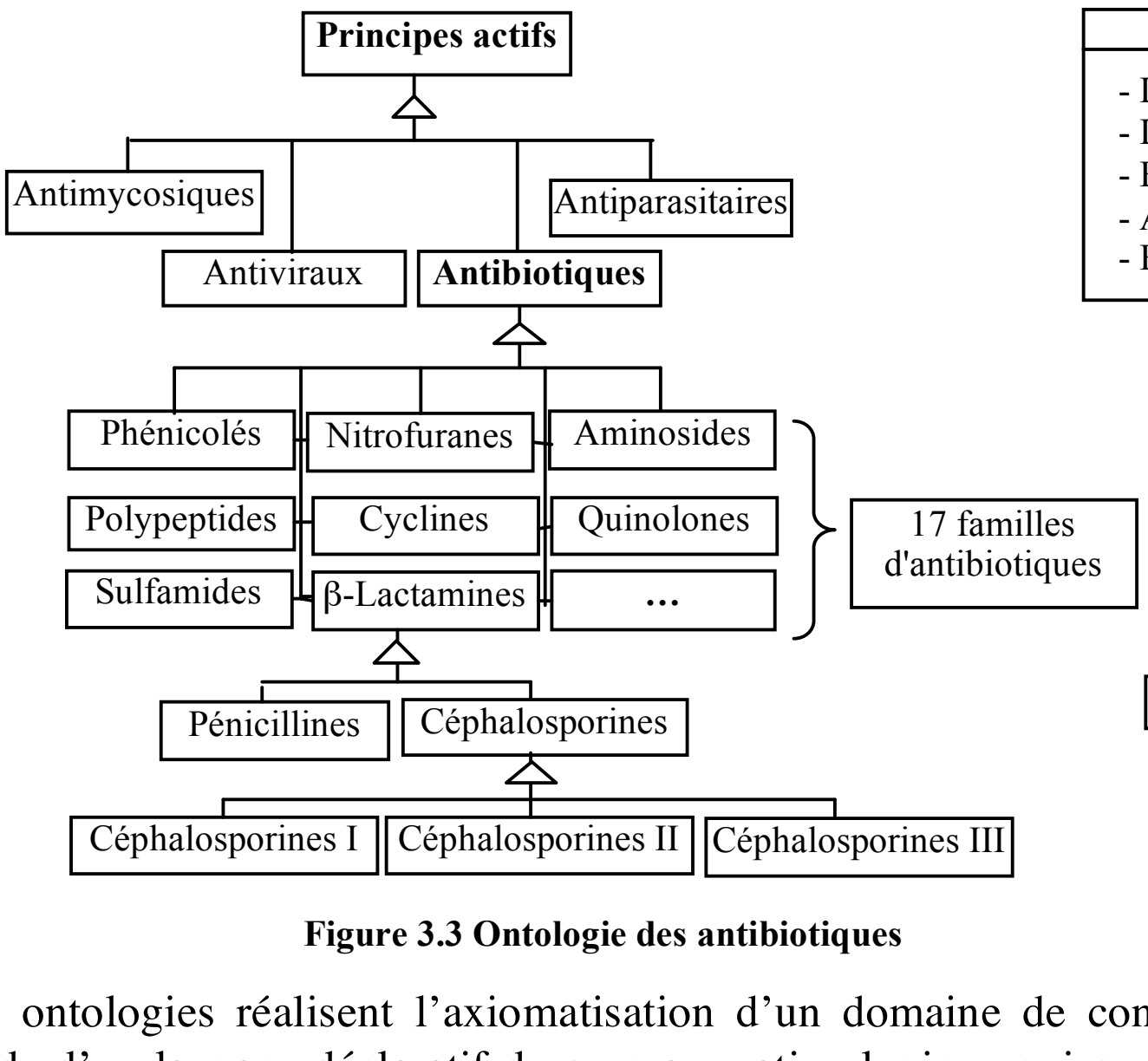

**Figure 3.3 Ontologie des antibiotiques**

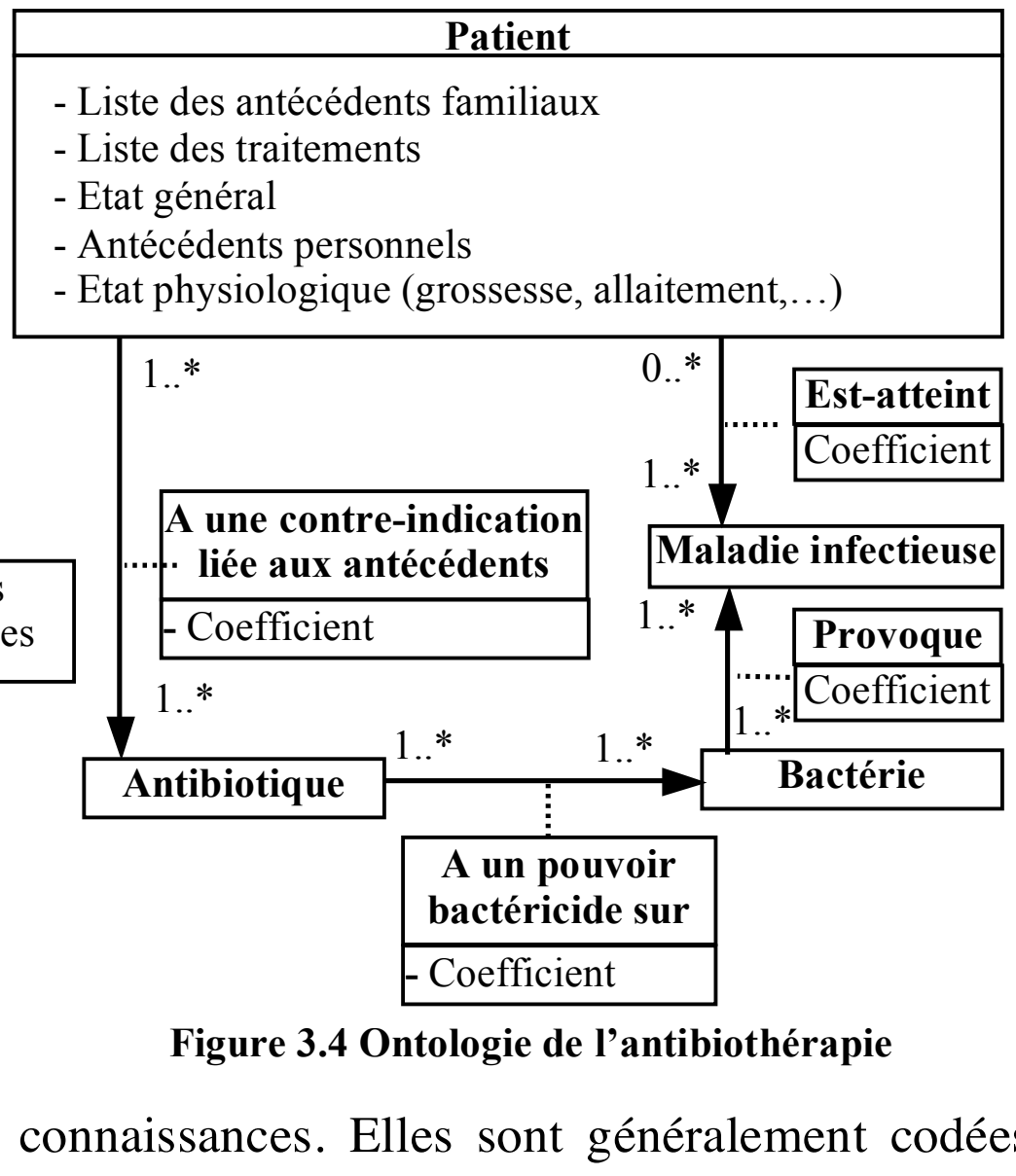

**Figure 3.4 Ontologie de l'antibiothérapie**

Les ontologies réalisent l'axiomatisation d'un domaine de connaissances. Elles sont généralement codées à l'aide d'un langage déclaratif de programmation logique qui permet d'associer des descriptions écrites dans un langage commun aux différents agents avec des instructions formelles.

Notre modèle ontologique se fonde sur la richesse sémantique des modèles orientés objet. La notation UML (Unified Modeling Language) permet d'exprimer les relations sémantiques définissant la terminologie d'un domaine de connaissance [11,12]. En effet, sur les figures 3.1 à 3.4, nous montrons qu'un diagramme de classes UML offre des capacités sémantiques suffisantes, rigoureuses, graphiques pour coder une ontologie de l'antibiothérapie fondée sur les connaissances de SIAMED 2 (Système Informatique d'Antibiothérapie MEDicale) [5]. La figure 3.4 montre qu'une maladie bactérienne est provoquée par une ou plusieurs bactéries. L'ontologie de la figure 3.3 décrit la classification des antibiotiques. La figure 3.1 montre la classification des agents infectieux. Sur la figure 3.2, l'ontologie décrit les maladies comme une succession de tableaux cliniques comportant des signes cliniques pathognomoniques (**SP**), obligatoires (**SO**), évocateurs (**SE**) ou accessoires (**SA**) liés à une pathologie. ***Le type d'agent ontologique*** crée une seule instance d'agent ontologique par étape clinique d'un domaine de connaissance qui communique la terminologie commune à l'ensemble des agents spécialisés concernés par cette étape, par exemple : diagnostic des maladies infectieuses, prescription des antibiotiques. Les ontologies sont des représentations symboliques conceptuelles collectives sur lesquelles vont s'articuler les interactions entre agents spécialisés qui contribuent à l'élaboration d'une solution au problème commun posé par chaque étape clinique.

### 4. Architecture d'un système multi-agent d'aide à la décision (SMAAD) clinique

Le SMAAD instancie les types d'agents TAS, TACG, TAMC et TASDC au cours de deux étapes de spécialisation (figure 4). Durant la première spécialisation, les Types d'Agents Modèles de connaissances (TAMC) héritent des modules et fonctionnalités du TACG (figure 2) et sont dotés de fonctionnalités spécifiques d'un modèle des connaissances. Par exemple, un TAMC à base de règles aura accès à un moteur d'inférence. Un TAMC épidémiologique permettra de connaître l'incidence et la prévalence d'une pathologie. Un seul agent superviseur est instancié. Un AST est un Agent spécialisé dans une tâche clinique Δ, Π, Θ, SΘ.

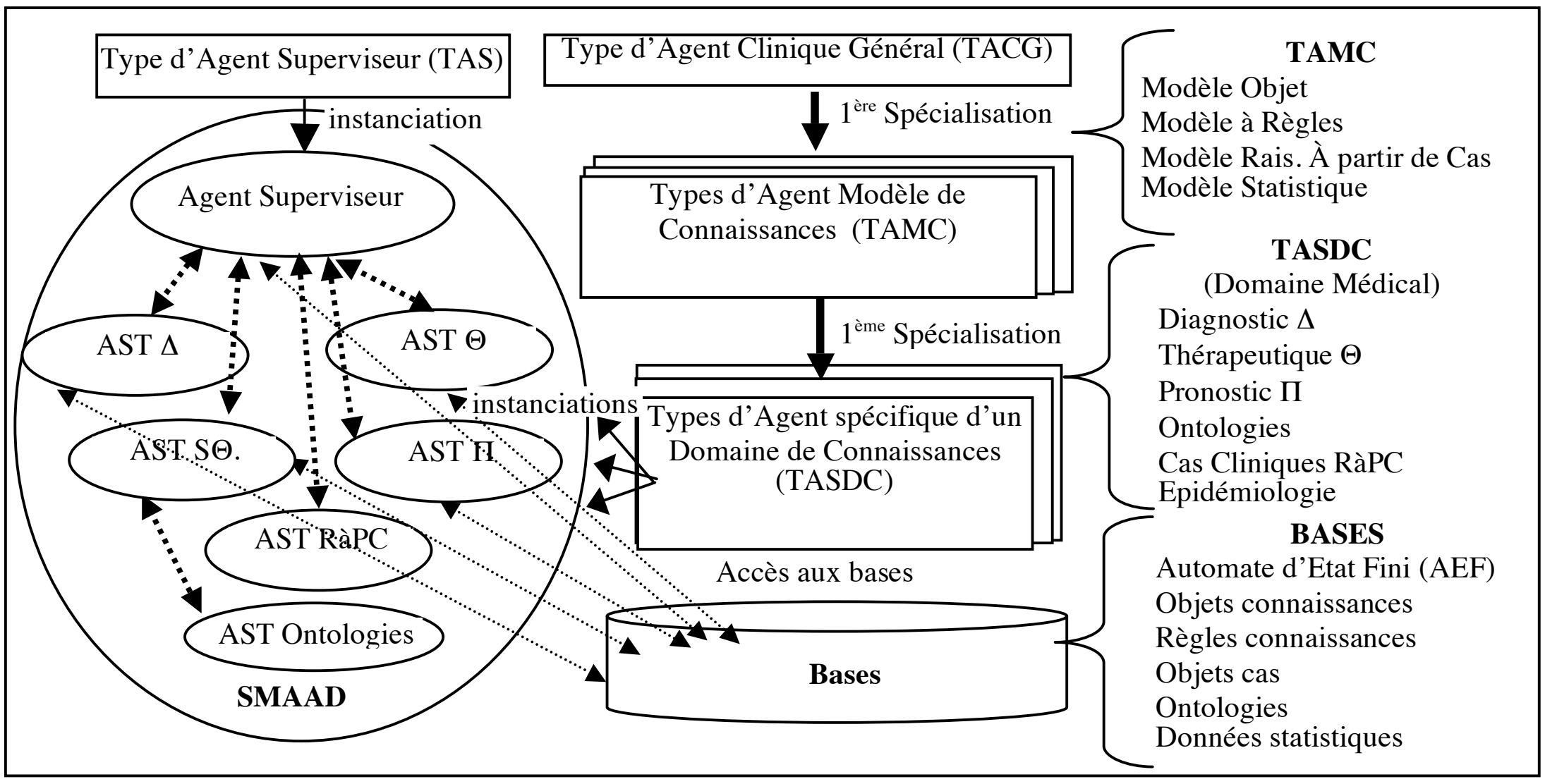

**Figure 4 : Spécialisation des agents impliqués dans un système multi-agent d'aide à la décision SMAAD**

La deuxième spécialisation produit des Types d'Agents spécifiques du Domaine de Connaissances (TASDC) et des étapes cliniques spécifiques Δ, Π, Θ, SΘ, par exemple : le ***diagnostic*** de maladies infectieuses, le ***pronostic*** de cette maladie selon sa sévérité et l'état général du patient, le ***traitement*** d'une maladie bactérienne par antibiothérapie. Ces agents utilisent des modèles de raisonnement et de connaissances hérités des TAMC (des règles de production, des distances entre objets connaissances, des objets cas cliniques, des statistiques etc.). Au cours de chaque étape clinique, les agents adressent des requêtes à des bases de connaissances spécifiques.

## 5. Un exemple : Diagnostic, pronostic et traitement d'une Diarrhée aiguë

La figure 5 ci-dessous illustre la démarche clinique implantée par le SMAAD. L'automate clinique général du superviseur (AEFCG) charge l'AEFΔ destiné au diagnostic d'une diarrhée aiguë. La présence initiale d'une diarrhée aiguë (SO1) peut conduire aux diagnostics Δ1 à Δ8 selon les signes cliniques SE1 à SE6, les antécédents cliniques (AC) et les signes de gravités (SG) constatés chez le patient par l'utilisateur et selon les transitions de l'automate (figure 5.2.). Ensuite, l'AEF Π, définit les pronostics associés : Π1, Π3, Π5 à Π8, puis l'AEF Θ définit les stratégies thérapeutiques correspondantes : Θ1, Θ3, Θ5 à Θ8 (tableau 5.3). Les suivis thérapeutiques SΘ, ne sont pas présentés par concision. Le tableau 5.4. décrit trois cas cliniques avec les transitions déclenchées. L'AEF RàPC (figure 1) stocke et indexe successivement ces cas et les liens de composition de leurs composants dans la base de cas (figure 4) avec les éléments (Δ, Π, Θ, SΘ) avec les mots clés définissant le problème du cas (PB), l'environnement (fiche du patient) (E), le résultat obtenu (R). Ces cas enrichissent la base de cas en vue d'une nouvelle utilisation du SMAAD dans ce contexte [7,14,19].

| Tableau 5.1 Légende Signes cliniques obligatoires (SO), évocateurs (SE) | |
|---|---|
| $SO_1$ | Selles fréquentes récentes molles ou liquides |
| $SE_1$ | Etat nauséeux ou vomissement, déshydratation extra-cellulaire, glaire, sang, douleurs abdominales (ténesme, épreintes), fièvre, fatigue, myalgies, contage, voyage zone tropicale, récidive, durée > 5jours |
| $SE_2$ | résultat coproculture positif et antibiogramme |
| $SE_3$ | résultat parasitologie selles + |
| $SE_4$ | bilan colique : Abdomen sans préparation, coloscopie |
| $SE_5$ | hémoculture |
| $SE_6$ | Recherche de toxiques + |
| AC | **Antécédents à risque** : immunodépression, insuffisance rénale, valvulopathie, cirrhose, diabète, SIDA |
| SG | **Signes de Gravité** : rectorragies, glaire, fièvre >= 38°5 ou hypothermie, accès bactériémiques, collapsus |

**Figure 5.2 diagnostic, pronostic d'une diarrhée aiguë**

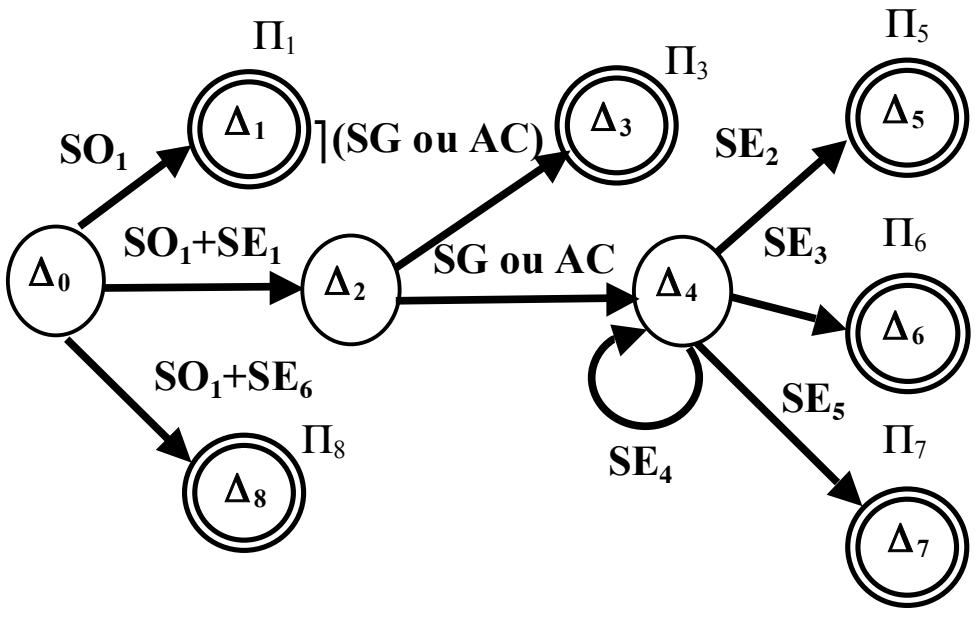

| Tableau 5.3 Diagnostic, pronostic et traitement d'une diarrhée aiguë | | | | | |
|---|---|---|---|---|---|
| Diagnostics | | Pronostics | | Thérapie | |
| $\Delta_0$ | Diarrhée aiguë | | | | |
| $\Delta_1$ | Diarrhée virale (Rotavirus, Parvovirus…) | $\Pi_1$ | Bénin | $\Theta_1$ | hydratation orale, antikinétiques opiacés (codeine, diphénoxylate, lopéramide), smectites+ régime |
| $\Delta_2$ | Diarrhée à germe invasif | | | | |
| $\Delta_3$ | Diarrhée bactérienne bénigne | $\Pi_3$ | Curable | $\Theta_3$ | $\Theta_3$ hydratation orale ou IV, chélateurs des toxines (colestyramine Questran®)), coproculture, antibiogramme et antibiothérapie adaptée au germe |
| $\Delta_4$ | Diarrhée bactérienne ou parasitaire sévère | | | | |
| $\Delta_5$ | Diarrhée bactérienne sévère | $\Pi_5$ à $\Pi_8$ | Vital | $\Theta_5$ | hospitalisation, $\Theta_3$ + correction hypovolémie (plasma+albumine),antibiothérapie IV. Clostridium difficile (Vanomycine : Vancocine®)) |
| $\Delta_6$ | Diarrhée parasitaire sévère | | | $\Theta_6$ | idem à $\Theta_5$ + recherche parasitaire, antiparasitaire : métronidazole (Flagyl®) ou Tinidazole (Fasigyne®) puis Tiliquinol (Intetrix®). |
| $\Delta_7$ | Septicémie (germes gram -) | | | $\Theta_7$ | idem à $\Theta_5$ avec hémoculture + antibiothérapie IV |
| $\Delta_8$ | Diarrhée toxique | | | $\Theta_8$ | Recherche du toxique+ chélateur du toxique |

**Tableau 5.4 Cas cliniques (RàPC)**

**Cas clinique 1** Un homme de 45 ans consulte pour une diarrhée d'apparition brutale avec 5 exonérations fécales liquides ces derniers jours. L'examen est normal sans signe de déshydratation, l'ionogramme sanguin est normal, la température à 37,8°C. Ce cas est $\Delta_1$ , $\Pi_1$ , $\Theta_1$ et S$\Theta$ : Consulter à nouveau si persistance > 5 jours ou fièvre > 38°5.

**Cas clinique 2** Depuis 3 jours un homme de 28 ans après un voyage en Sardaigne se plaint d'une diarrhée acqueuse abondante avec 10 exonérations quotidiennes diurnes et nocturnes. Les vomissements rendant difficile l'alimentation. Le soir, le fièvre est soir à 38°5. La palpation montre un abdomen sensible mais souple, pas de déshydratation. L'ionogramme sanguin est normal. Ce cas est $\Delta_2$ confirmé en $\Delta_3$ par la coproculture : salmonellose mineure sensible. Ce cas est $\Pi_3$ $\Theta_3$ antibiothérapie avec Amoxicilline (Clamoxyl®) IV 4g /jour. S$\Theta$ : surveillance de la diurèse et du ionogramme.

**Cas clinique 3** Depuis 8 jours un homme de 60 ans se plaint de troubles du transit : plus de 10 selles par jour liquides, une fièvre > 39°C, un météorisme abdominal, une déshydratation avec une pression artérielle à 9/6, une oligurie. On retrouve un traitement par ampicilline. La coloscopie évoque une colite pseudo-membraneuse et la coproculture met en évidence clostridium difficile. Ce cas est $\Delta_5$ , $\Pi_5$ , $\Theta_5$ avec hospitalisation et Vancocine ® 500mg 4/jour soit 2g/j et S$\Theta$ surveillance de la diurèse et de la volémie.

**Figure 5. Diagnostic, pronostic et traitement d'une Diarrhée aïgue**

# 6. Comparaison avec d'autres approches

L'utilisation des systèmes multi-agents pour la coopération de bases de connaissances n'est pas récente et a donné lieu à de nombreux travaux, par exemple [4, 10, 16, 17]. Le SMAAD proposé dans cet article assure la coopération de composants connaissances cliniques hétérogènes. Il s'appuie sur un méta-modèle de démarche clinique et des travaux menés depuis une quinzaine d'années [6,7,19]. Il offre une approche interactive avec l'utilisateur via l'agent superviseur. Le SMAAD assure l'interopérabilité sémantique des données et des connaissances par les agents ontologies couplées aux agents connaissances. SAPHIRE [10] est un système plus macroscopique, fondé sur des directives cliniques et des agents ontologiques. Il permet l'interopérabilité d'appareils, de bases de données et de connaissances. Il n'est pas guidé par un modèle de démarche clinique. Notre SMAAD ne gère pas l'accès par plusieurs utilisateurs dans un workflow clinique comme dans [4,13].

# 7. Conclusion

Nous avons présenté une architecture multi-agents (SMAAD) permettant la prise en compte de l'ensemble du cycle de décision clinique adaptable à de nombreux domaines médicaux. Cette approche se fonde sur la spécialisation d'agents par modèle et par domaines de connaissances convoqués aux étapes cliniques (le diagnostic, le pronostic, le traitement, le suivi thérapeutique). Les agents spécialisés assurent l'interopérabilité et la coopération des bases connaissances encapsulées. Ils partagent une terminologie commune enrichie de liens sémantiques grâce à des agents ontologies. Notre approche est adaptable à de nombreuses spécialités médicales mais aussi à d'autres domaines où une démarche clinique est pertinente (par exemple : la gestion d'entreprise, la psychologie clinique, les cultures agricoles, la botanique, le développement durable …).

## Références